\documentclass[11pt,a4paper]{article}
\usepackage[hyperref]{ranlp2023}
\usepackage{times}
\usepackage{latexsym}

\usepackage{amsmath, amssymb}
\usepackage{enumitem}
\usepackage{geometry}
\usepackage{microtype}
\usepackage{array}
\usepackage{graphicx}

\usepackage{booktabs}
\usepackage{multirow}

\usepackage{microtype}

\aclfinalcopy 

\title{Prompt Balance Matters: Understanding How Imbalanced Few-Shot Learning Affects Multilingual Sense Disambiguation in LLMs}

\author{
    Deshan Sumanathilaka, Nicholas Micallef, Julian Hough \\
    Department of Computer Science, Swansea University, Wales, UK \\
    { \{t.g.d.sumanathilaka, nicholas.micallef, julian.hough\}@swansea.ac.uk}
}

\date{}

\begin{document}
\maketitle
\begin{abstract}
Recent advances in Large Language Models (LLMs) have significantly reshaped the landscape of Natural Language Processing (NLP). Among the various prompting techniques, few-shot prompting has gained considerable attention for its practicality and effectiveness. This study investigates how few-shot prompting strategies impact the Word Sense Disambiguation (WSD) task, particularly focusing on the biases introduced by imbalanced sample distributions. We use the GLOSSGPT prompting method, an advanced approach for English WSD, to test its effectiveness across five languages: English, German, Spanish, French, and Italian. Our results show that imbalanced few-shot examples can cause incorrect sense predictions in multilingual languages, but this issue does not appear in English. To assess model behavior, we evaluate both the GPT-4o and LLaMA-3.1-70B models and the results highlight the sensitivity of multilingual WSD to sample distribution in few-shot settings, emphasizing the need for balanced and representative prompting strategies. 
\end{abstract}

\section{Introduction}

With the advent and rapid development of transformer architectures, Large Language Models (LLMs) have emerged as a game-changing technology for Natural Language Processing (NLP) tasks, particularly in text generation, question answering, and tasks requiring computational intelligence, reasoning, and language understanding~\cite{minaee_large_2024}. Previous research has explored a variety of computational techniques in relation to LLMs, with a strong focus on prompt engineering, Retrieval Augmented Generation (RAG), knowledge base integration, and efficient fine-tuning strategies~\cite{gu2024survey}.

Among these areas, prompt engineering has received significant attention as a means of constructing accurate and efficient responses. Notably, few-shot prompting~\cite{mann2020language} has been extensively studied to enhance reasoning capabilities and in-context learning within prompting strategies.

Recent work, such as GLOSSGPT\footnote{https://github.com/Sumanathilaka/GlossGPT-GPT-4-WSD-with-COT}, has achieved state-of-the-art performance on the WSD task in English by leveraging few-shot prompting strategies. This approach demonstrates a strong ability to resolve lexical ambiguity~\cite{sumanathilaka_glossgpt_2025}. Other work has shown that zero-shot prompting alone cannot perform efficient WSD, but few-shot chain-of-thought (COT) can lead to higher-accuracy disambiguation~\cite{sumanathilaka2024assessing}. WSD remains a critical computational challenge for improving the understanding of word meanings when ambiguous terms appear in sentences or paragraphs. Effective WSD systems also contribute indirectly to advances in computational translation, transliteration, question answering and language understanding. While GLOSSGPT has demonstrated strong effectiveness for English, its generalizability to other languages remains unexplored. This research aims to address that gap by investigating whether the same approach can be effectively applied in a multilingual setup. In doing so, the study also examines how few-shot prompting may introduce bias into classification tasks such as WSD, specifically exploring whether models tend to favor high-frequency senses over low-frequency ones\footnote{Senses that are uncommon or rarely used}. To analyze this behavior, we employ three sampling techniques namely Highest Frequency Sharing, Lowest Frequency Sharing, and Average Frequency Sharing as detailed in Section~\ref{sec:Methodology}. Our findings and discussions are presented accordingly.

This study makes the following major contributions:

\begin{itemize}
    \item We systematically investigate how different few-shot sampling strategies (Highest, Lowest, and Average Frequency Sharing) influence WSD performance across five languages. Our multilingual setup reveals that sense frequency imbalance introduces varying degrees of bias, with under-resourced languages being especially vulnerable.
    \item Our findings further highlight the importance of maintaining balanced few-shot examples as a critical factor for mitigating bias and improving disambiguation accuracy, especially in low-resource language contexts.
    \item We demonstrate that the optimal prompting strategy is language and model-specific, showing that a one-size-fits-all prompting approach fails to generalize effectively.
\end{itemize}

The remainder of the paper is organized as follows: Related Work, which discusses current research on multilingual WSD and few-shot bias studies; Methodology, which outlines the approach used to evaluate our proposed study; Results and Observations; and finally, Conclusions and Future Directions, which address potential strategies to mitigate bias in classification tasks across different language settings.

\section{Related Work}
\label{sec:Relatedworks}
 
We divide this section into two subsections. The first subsection describes WSD experiments in the context of Language Models (LMs) and LLMs, including recent advances. The second section describes experiments related to few-shot bias detection.

\begin{figure*}[h!]
    \centering
    \includegraphics[width=0.19\textwidth]{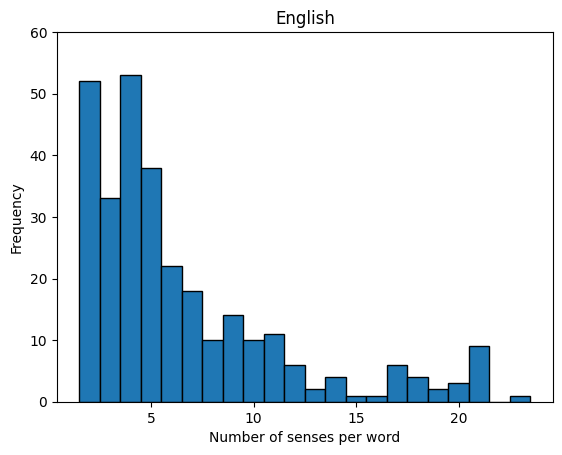}
    \includegraphics[width=0.19\textwidth]{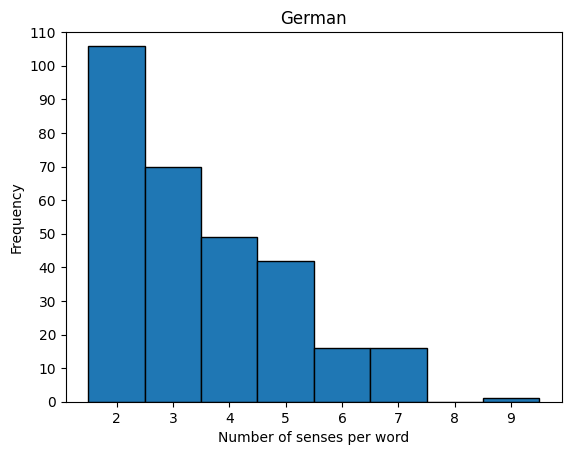}
    \includegraphics[width=0.19\textwidth]{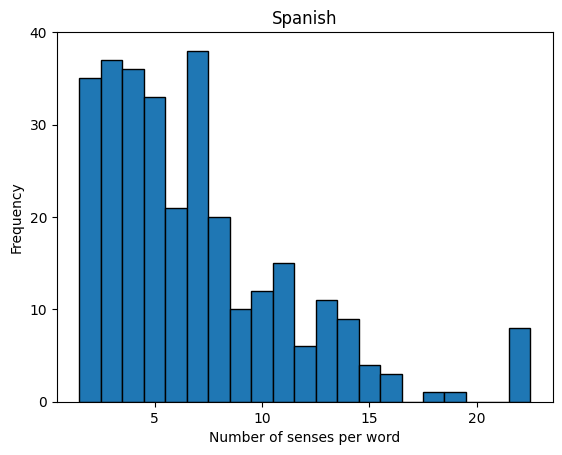}
    \includegraphics[width=0.19\textwidth]{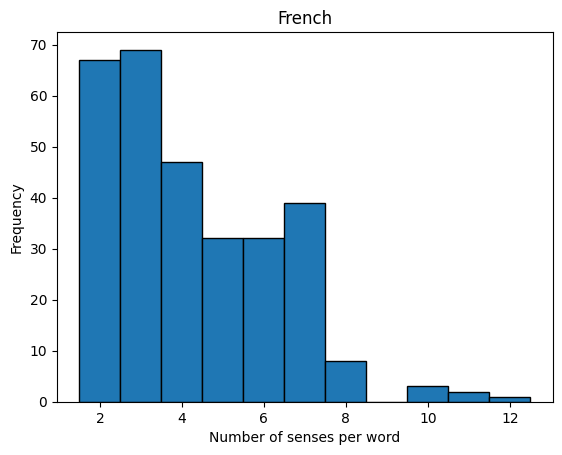}
    \includegraphics[width=0.19\textwidth]{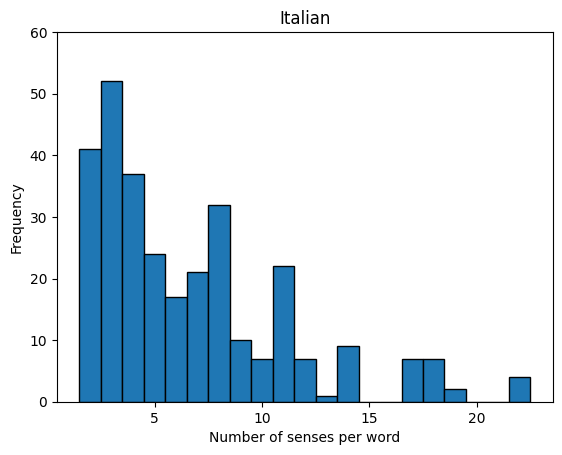}
    \caption{Sense distribution for selected samples on each language. The order is English, German, Spanish, French and Italian }
    \label{senseDistribution}
\end{figure*}

\subsection{Advancements in Language models for WSD}

Recent developments in language models have generated substantial interest in evaluating their performance across a range of NLP tasks.~\citet{sainz_what_2023} demonstrated that LLMs possess an inherent ability to capture word senses, indicating their potential for WSD without explicit task-specific training. They framed WSD as a textual entailment task, prompting LLMs to assess the appropriateness of a domain label for a sentence containing an ambiguous word. Notably, this zero-shot approach surpassed random baselines and, in certain cases, matched or even outperformed supervised WSD systems~\cite{ortega-martin_linguistic_2023}. Additionally, cross-lingual WSD has been explored through contextual word-level translation using pre-trained language models, with evaluations of zero-shot performance based on cross-lingual knowledge~\cite{kang_translate_2023}. A contrastive self-training framework, COSINE, was also proposed to fine-tune pre-trained LLMs using weak supervision without requiring labeled data~\cite{yu_fine-tuning_2021}.~\citet{manjavacas2022non} investigated non-parametric learning approaches and fine-tuning strategies for LLMs applied to historical Dutch and English corpora.~\citet{qorib2024decoder} highlighted the comparative effectiveness of encoder-only models over decoder-only architectures.~\citet{yae2024leveraging} examined the impact of LLM size on WSD performance, while~\citet{cahyawijaya2024thank} revealed limitations in cross-lingual WSD tasks, particularly involving false friends.\footnote{Orthographically similar words that have entirely different meanings across languages}

Furthermore,~\citet{sumanathilaka2024assessing} demonstrated that prompt engineering techniques can significantly enhance WSD performance through in-context learning using GPT-3.5 Turbo and GPT-4-turbo. Their study explored various prompting strategies, including zero-shot, few-shot, and few-shot-CoT, highlighting the effectiveness of few-shot learning in improving sense prediction accuracy.  It also showed that incorporating external knowledge further enhances the effectiveness of sense disambiguation. This work was further extended in subsequent studies, which showed that models such as Deepseek-R1 and o4-mini performed particularly well in WSD tasks compared to other flagship LLMs~\cite{sumanathilaka2024can}. These findings are also supported by~\citet{kibria2024functional}. A key source of inspiration for this line of research is \textit{GLOSSGPT}~\cite{sumanathilaka_glossgpt_2025}, which achieved state-of-the-art performance in English WSD by leveraging knowledge base-driven few-shot prompting. The model effectively incorporated lexical knowledge using WordNet glosses and synonyms. Although this approach outperforms several existing WSD systems, the direct impact of few-shot learning requires further investigation. To address this, we propose a sampling-based approach aimed at gaining deeper insights into how various few-shot configurations influence WSD performance across languages.

\subsection{Investigating Bias in Few-Shot Learning with LLMs}

A very few recent studies have focused on the biases introduced by few-shot prompting in classification tasks, particularly in contexts involving LLMs. These biases often stem from prompt design, example selection, and label distribution, and can significantly affect model fairness and performance consistency.

The study by~\citet{lai2025sas} introduces a benchmark specifically for assessing short answer scoring with few-shot prompting. It highlights how LLMs amplify biases when prompted with limited, unbalanced examples and shows how model predictions become skewed toward overrepresented classes.~\citet{mallen2024balancing} analyze the trade-off between label quantity and quality in few-shot prompts. These experiments reveal that weak labels often introduce substantial bias, especially in binary classification tasks. They also highlight that using a combination of low-quality and high-quality labels has a positive impact on the prediction process rather than either alone.

The study by~\citet{ma2023fairness} revisits the problem of predictive bias, introducing a novel evaluation metric and proposing two algorithms namely T-fair Prompting and G-fair Prompting that aim to improve classification performance by selecting support examples that yield a more uniform distribution over output classes. More recently,~\citet{ahmadnia2025active} emphasized that Few-Shot Learning performance degrades significantly when inappropriate support samples are selected. To address this, they introduced a new method that combines fine-tuning with Active Learning (AL) for support sample selection. Their approach leverages embedding techniques to extract salient features from unlabeled data and applies strategic sampling to select the most informative examples, thereby enhancing classification outcomes. Similarly,~\citet{pecher2024automatic} highlighted the crucial role of sample quantity and quality in few-shot learning. Their work investigates how different sample selection strategies can be combined to mitigate the limitations posed by a restricted number of training examples and improve overall learning effectiveness. 

These studies underscore the critical impact of few-shot prompting strategies on classification tasks, particularly emphasizing how imbalanced sample distributions can introduce predictive biases and affect both fairness and model reliability.

\section{Methodology}
\label{sec:Methodology}

\begin{figure*}[t]
  \centering
  \includegraphics[width=\textwidth]{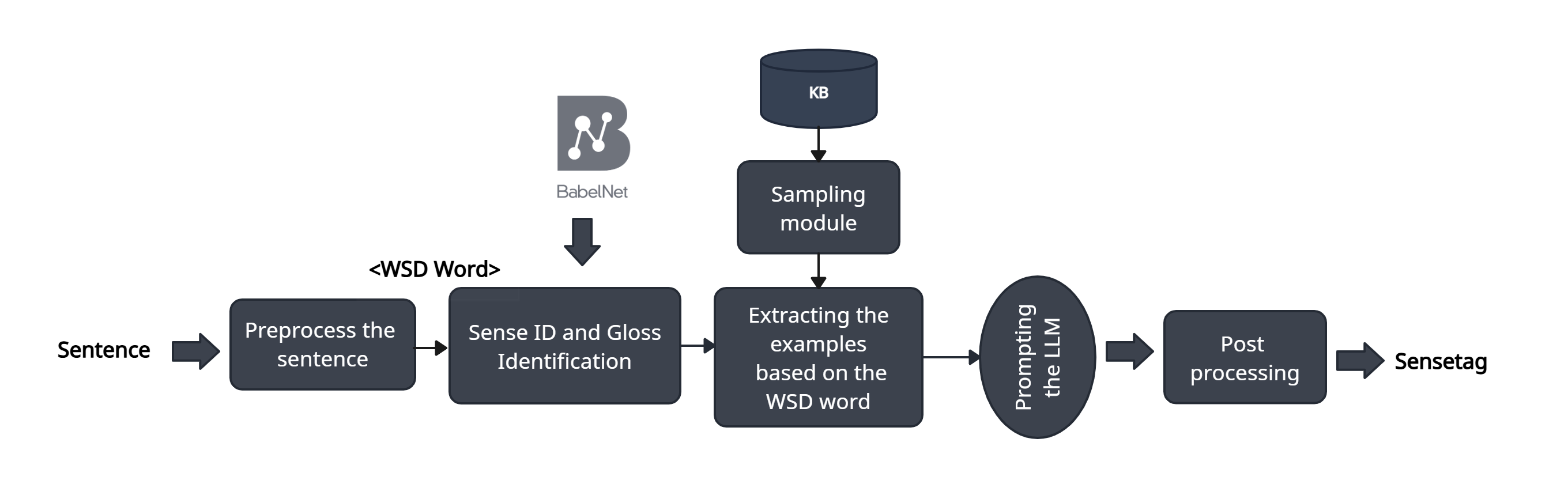}
  \caption{The data flow of the experiment process.}
  \label{fig:flow}
\end{figure*}

This study has built upon a previously verified few-shot COT prompt provided by the GLOSSGPT. Prompts have been designed in English following a systematic chain of thought process, sequentially providing the lexical resources (gloss + synonyms) and a few possible few-shot instances extracted from pre-built KB, as illustrated in Figure~\ref{fig:flow}. English prompts were used in all experiments, including multilingual setups, to minimize prompt ambiguity during inference. This ensures that the core evaluation remains focused on the WSD task itself, rather than being influenced by prompt design \cite{aina2021language}. In this section, we present the dataset used, the techniques employed for knowledge base creation, and the sampling method applied for frequency sharing. 

\subsection{Datasets}

\subsubsection{SemEval-2013 WSD dataset}

For our evaluation, we use an updated version of SemEval-13\footnote{https://github.com/SapienzaNLP/mwsd-datasets}, which contains four languages: Italian (IT), Spanish (ES), French (FR), and German (DE). English WSD was evaluated using the SemEval-13 English dataset~\cite{jurgens2013semeval}.

The sentence with multiple ambiguous words is split into different sentences, ensuring that each sentence contains only one ambiguous word, which is enclosed between \texttt{<WSD>} tokens for the LLM inference task. A total of 300 random samples are utilized for the study for each language, with each sentence containing exactly one ambiguous word marked for disambiguation. A total of 300 samples were carefully selected to ensure that each ambiguous word had at least two distinct senses in BabelNet, a prerequisite for meaningful disambiguation. Priority was given to nouns, although samples also included other parts of speech (POS). The polysemy histogram for each language is shown in Figure~\ref{senseDistribution}. Furthermore, the study is constrained to a limited number of samples due to the practical limitations imposed by BabelNet’s inference capabilities via API. To ensure consistency across evaluations, the same set of 300 random samples is used across all three sampling methods discussed in subsection~\ref{sampling}. The micro F1 score is employed as the evaluation metric for assessing model performance.

\subsubsection{BabelNet}

BabelNet~\cite{navigli2010babelnet} is the primary lexical knowledge base used for this study. It is a multilingual lexical and encyclopedic resource built by semi-automatically integrating various sources such as WordNet, multilingual WordNets, and Wikipedia. It contains multilingual synsets of synonymous terms across different languages, spanning 600 languages and includes over 23 million synsets. For this study, lexical knowledge resources are primarily obtained through the BabelNet API, which imposes a daily limit of 1,000 BabelCoins. This constraint necessitated limiting the study to a smaller set of samples. In addition, English lemmas and their corresponding synonyms are extracted from WordNet~\cite{miller1995wordnet} to further enrich the lexical representation, especially in capturing and disambiguating ambiguous word meanings.

\subsection{Knowledge-base creation for few-shot retrieval}
\begin{figure}[t]
  \includegraphics[width=\columnwidth]{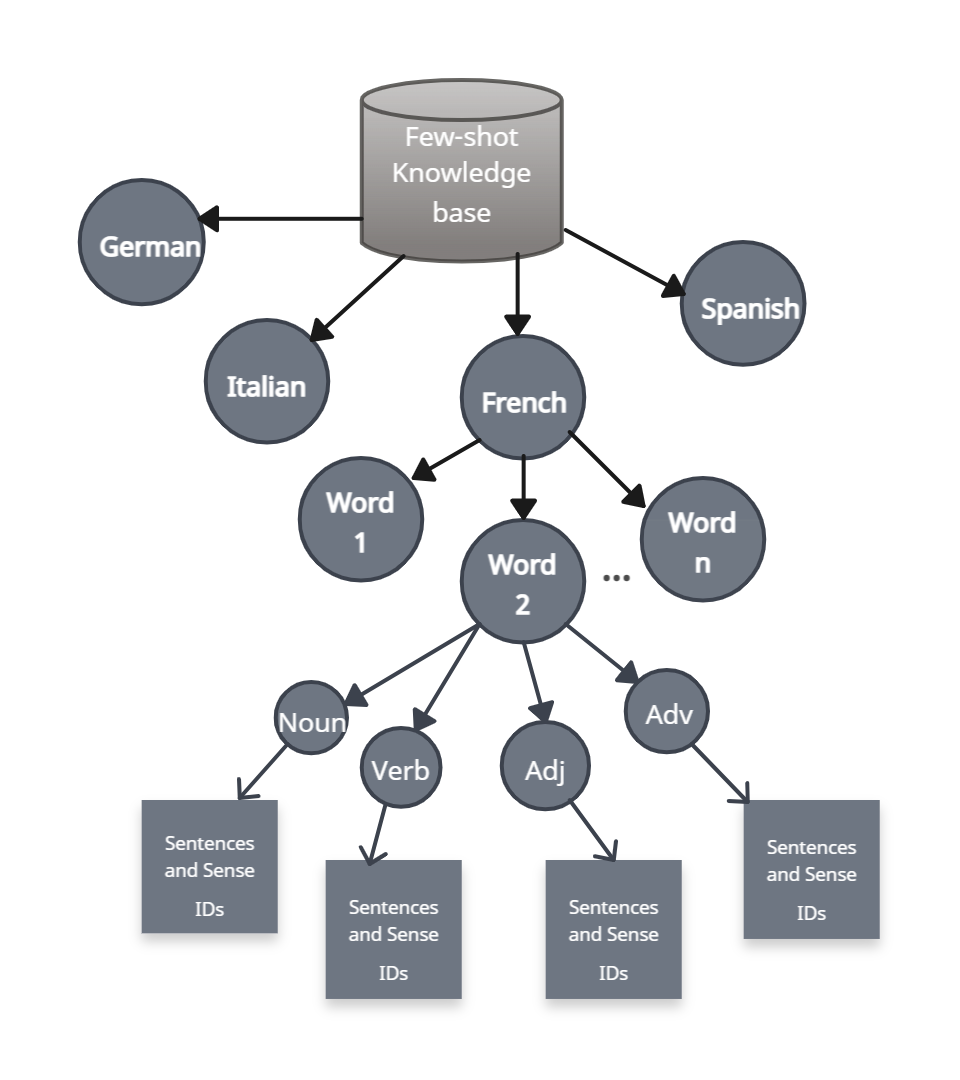}
  \caption{Few-shot knowledge base arrangement. For demonstration, the French branch is shown. A similar arrangement is followed for German, Italian and Spanish.}
  \label{fig:kbgraph}
\end{figure}

The creation of the knowledge base (KB) was inspired by GLOSSGPT~\cite{sumanathilaka_glossgpt_2025} and has been further enhanced to support a multilingual setup. The training data for all four languages is structured as a tree, with the language as the root node. The first-level parent nodes represent ambiguous words, the second-level nodes correspond to POS tags, and the child nodes contain example instances along with their respective BabelNet sense IDs. For efficient retrieval, the structure is stored in a JSON file. Based on the ambiguous word, the required information can be retrieved in constant time and shared with the model for few-shot prompting, following the sampling strategies described in subsection~\ref{sampling}. A detailed structure is provided in Figure~\ref{fig:kbgraph}.

\subsection{Sampling Strategies for Few-Shot Prompting in WSD}
\label{sampling}

\begin{table*}[ht]
\centering
\begin{tabular}{|l|c|c|c|c|c|}
\hline
\textbf{Word Sense} & \textbf{Meaning Description} & \textbf{Actual \#} & \textbf{HF}&\textbf{LF}&\textbf{AF}\\
\hline
\texttt{bank.n.14:00} & Financial institution & 7 & 7 & 1 & 4 \\
\hline
\texttt{bank.n.17:01} & Edge/slope of a river or body of water & 4 &4&1&4 \\
\hline
\texttt{bank.n.17:00} & Raised embankment, like a ridge or mound & 1 & 1 & 1 & 1\\
\hline
\texttt{bank.n.14:01} & Series/set (e.g., a bank of windows) &1 & 1 & 1 & 1 \\
\hline 
\end{tabular}
\caption{Sense Count and Example Report for the Word ``Bank'', according to the three frequency sharing techniques. H: High, L: Low, A: Average, F: Frequency}
\label{tab:bank-sense-count}
\end{table*}

In this study, we apply few-shot prompting using the in-context learning paradigm to identify the correct sense of an ambiguous word in the WSD task. We explore how the frequency distribution of senses in the example pool affects the model's performance. Specifically, we define three sampling strategies based on the distribution of sense frequencies below: \textit{Highest Frequency Sharing}, \textit{Lowest Frequency Sharing}, and \textit{Average Frequency Sharing}. We denote:

\begin{itemize}[label=--]
    \item $n$ as the total number of senses for a given ambiguous word,
    \item $\text{freq}(S_i)$ as the number of available examples for sense $S_i$, where $i \in \{1, 2, ..., n\}$,
    \item $\mathcal{F} = \{\text{freq}(S_1), \text{freq}(S_2), \dots, \text{freq}(S_n)\}$ as the set of all sense frequencies.
\end{itemize}

Each strategy defines how many few-shot examples $k_i$ are selected for each sense $S_i$, where $k_i \leq \text{freq}(S_i)$. Case selection is performed randomly when $\text{freq}(S_i) > k_i$.

\subsection*{1. Highest Frequency Sharing}

This strategy aims to balance the number of few-shot examples according to the most frequent sense. Each sense $S_i$ is assigned $k = \max(\mathcal{F})$ examples, if sufficient samples are available, else the number of samples available for that particular sense:

\[
k_i = \min\left(\max(\mathcal{F}), \text{freq}(S_i)\right)
\]

\subsection*{2. Lowest Frequency Sharing}

This strategy equalizes the number of few-shot examples using the least frequent sense that has at least one sample. Each sense $S_i$ receives:

\[
k_i = \min\left(\min(\mathcal{F} \setminus \{0\}), \text{freq}(S_i)\right)
\]

This ensures that all senses are represented equally, without exceeding their available samples.

\subsection*{3. Average Frequency Sharing}

This strategy computes an average of the minimum non-zero and maximum frequencies and uses it as a balanced number of examples per sense:

\[
k = \left\lfloor \frac{\min(\mathcal{F} \setminus \{0\}) + \max(\mathcal{F})}{2} \right\rfloor
\]
\[
k_i = \min(k, \text{freq}(S_i))
\]

This method serves as a compromise between the two extremes and reduces the effect of extreme imbalance.

Occasionally, it is observed that certain senses have no corresponding examples in the training data used to build the knowledge base for few-shot retrieval. In such cases, regardless of the sampling strategy applied, only the sense identifier (sense ID) is shared in the prompt, without any supporting examples. The frequencies and adjusted frequencies according to the three frequency sharing techniques is shown for \textit{``Bank''} as a noun in Table~\ref{tab:bank-sense-count}.

\subsection{Study Setup}

For this study, we selected the GPT-4o and LLaMA 3.1-70B models due to their strong performance in multilingual settings \cite{vayani2025all}. General-purpose chat models were used in this study without any fine-tuning or prompt tuning. This was deliberately done to ensure the evaluation of the effectiveness of prompt engineering alone, using a few-shot example. Access to GPT-4o was obtained via the OpenAI API using a tier-one OpenAI account, while LLaMA-3.1-70B was accessed through the Together.ai API. Both models were configured with a temperature of 0 and a maximum output token limit of 500. The temperature selection for the study was inspired by previous work~\cite{sumanathilaka2025exploring}, and zero(0) was selected to ensure the deterministic responses of the study. The primary task assigned to both LLMs was word sense identification, with their role defined as a ``\textit{helpful assistant for identifying word senses}''.

\begin{table*}[!t]
\centering
\begin{tabular}{llccccc}
\toprule
\textbf{Model} & \textbf{Method} & \textbf{English}& \textbf{German} & \textbf{Spanish} & \textbf{French} & \textbf{Italian} \\
\midrule
\multirow{3}{*}{GPT-4o} 
& Highest frequency & 0.81 & 0.76 & \textbf{0.70} & 0.75 & \textbf{0.74} \\
& Lowest frequency & 0.81 & 0.72 & 0.60 & 0.70 & 0.65 \\
& Average frequency  & \textbf{0.83}  & \textbf{0.78} & 0.64 & \textbf{0.76} & 0.70 \\

\midrule
\multirow{2}{*}{LLaMA-3.1} 
& Highest frequency & 0.75 & 0.76 & \textbf{0.73} & 0.72 & \textbf{0.68} \\
& Lowest frequency & \textbf{0.77} & 0.70 & 0.60 & 0.68 & 0.63 \\
& Average frequency  & 0.76 & \textbf{0.77} & 0.65 & \textbf{0.74} & 0.66 \\
\bottomrule
\end{tabular}
\caption{Performance comparison across languages for GPT 4o and LLaMA 3.1 under different frequency strategies. F1 scores are presented.}
\label{tab:frequency_comparison}
\end{table*}

\section{Results and Discussion}\label{sec:Results}
This section presents the results of our experiments, organized into three main areas of analysis. First, we evaluate the impact of different sampling strategies on performance. Next, we examine the suitability of various models for the task, highlighting their strengths and limitations. Finally, we explore the influence of contextual factors on the outcomes, providing insight into how context affects model behavior and overall system performance.

\subsection{Effectiveness of Sampling Strategies}

The experimental results in Table~\ref{tab:frequency_comparison} show there is no universally optimal few-shot sampling strategy for WSD. The efficacy of any given strategy is highly context-dependent. \textit{Average Frequency Sharing} often serves as a robust baseline, especially when paired with more capable models such as GPT-4o and applied to languages like English, German, and French in this study. Its balanced approach to sense representation generally proves beneficial, avoiding assigning too much weight to the high-frequency senses. \textit{Highest Frequency Sharing} emerges as a specialized but highly effective strategy in certain linguistic contexts, specifically Spanish and Italian in this dataset, where it consistently outperforms other methods for both LLMs. This suggests that high-frequency examples can improve sense prediction in lower-resourced languages like Spanish and Italian.

In contrast, \textit{Lowest Frequency Sharing} is generally a high-risk strategy, often resulting in suboptimal or even the worst performance. Its occasional success appears to be tied to specific model-language combinations, for example, LLaMA 3.1 with English. However, the performance improvement is not significant compared to the average frequency sampling. The variability in optimal strategy across different conditions, such as model and language, highlights the importance of empirical evaluation when aiming for peak performance. In practice, WSD applications should test multiple strategies or base their choice on strong, evidence-backed reasoning that considers the characteristics of the LLM and target language. This also suggests the potential value of developing adaptive methods that can dynamically select or adjust sampling strategies based on context and sense distribution.

Overall, these results suggest that frequency-based sampling has limited influence on WSD performance in English, where even the lowest frequency sense for a word has sufficient samples for the model to attune to, but plays a more significant role in multilingual contexts, particularly for less resourced languages like Spanish and Italian.

\subsection{Model-Specific Suitability for Few-Shot WSD}

The choice of LLM significantly affects the effectiveness of few-shot sampling strategies. In this WSD study, GPT-4o generally achieved the highest performance overall. However, LLaMA 3.1 proved to be a strong competitor, even outperforming GPT-4o in certain cases, such as when using \textit{Highest Frequency Sharing }in Spanish. Importantly, both models were sensitive to the sampling strategy used. Even a powerful model like GPT-4o did not perform best with a single strategy across all languages; for example, its performance in Spanish varied by 0.10 points depending on the sampling method. 

This shows that model size or general capability alone cannot fully compensate for poor sampling choices. On the other hand, a model like LLaMA 3.1 can deliver excellent results when the sampling method is well-matched to its strengths and the task at hand. Conversely, thoughtful sampling design can improve results even for smaller or less advanced models. These results suggest that future work could benefit from developing model-aware sampling techniques.

\subsection{Influence of Linguistic Context on Performance}

Linguistic context has a significant impact on WSD performance. This study revealed a consistent performance hierarchy across the five languages examined: English and German \textgreater French \textgreater  Spanish and Italian. This pattern held across both LLMs and most sampling strategies, indicating it reflects deeper linguistic or resource-based differences rather than specific methodological choices.

Importantly, the optimal sampling strategy also varied by language. \textit{Average Frequency Sharing} worked best for English, German, and French, while \textit{Highest Frequency Sharing} was more effective for Spanish and Italian. This suggests that under-resourced languages like Spanish and Italian benefit from sampling strategies that emphasize more frequent and balanced sense representations to improve interpretation and disambiguation. The key takeaway is the need for language-aware WSD strategies. Achieving strong multilingual performance requires more than powerful models, which demands careful attention to each language’s characteristics, including its sense distribution, resource availability, and representation in training data. This may involve tailored pre-processing, targeted resource development, or even fine-tuning models for specific languages or typological groups. A one-size-fits-all approach, typically optimized for English, is unlikely to perform well across the linguistic spectrum.

\section{Conclusion and Future Directions}

In conclusion, this research demonstrates that the selection of a few-shot examples in prompting LLMs can introduce significant performance variance in classification tasks in a multilingual setup, particularly when certain senses are overrepresented. However, in the case of English, such noticeable deviations are not identified. These results emphasize the importance of maintaining a balanced distribution of examples across all classes. The results also indicate that high-frequency sharing of sense examples can positively influence correct sense prediction, reinforcing the benefits of in-context learning during the inference process. Conversely, reducing few-shot examples to address class imbalance, especially for low-frequency senses, is not an effective strategy, as it can hinder the in-context learning process and degrade overall performance by limiting knowledge transfer. Imbalanced prompts tend to bias the model toward high-frequency senses, leading to reduced accuracy. While averaging techniques help mitigate such bias to some extent and contribute to more consistent performance, they are not a complete solution. 

Overall, these findings underscore the need for balanced few-shot prompting with sufficiently rich examples to teach LLMs accurate sense disambiguation. These insights are particularly valuable when extending similar techniques to low-resource languages, where inherent limitations in language performance make balanced prompting even more critical. In such low-resourced multilingual adaptation setups, ensuring a well-balanced distribution of examples can significantly enhance both in-context learning and classification accuracy.

Future studies should focus on methodologies for balancing and improving few-shot learning, particularly in low-frequency and uncommon-sense scenarios. As suggested by ~\citet{han2024llm, li2024survey}, multi-agent systems based on LLMs could be effectively utilized for context-aware few-shot generation, helping to create balanced examples necessary for the disambiguation process. These advancements can ensure that general-purpose LLMs are effectively leveraged for linguistic tasks such as WSD, rather than requiring fine-tuning for specific downstream applications.

The code and implementation are available at \url{https://github.com/Sumanathilaka/Prompt-Balance-Matters}.

\section*{Limitations}

One limitation of this study is that it considers only two flagship LLMs, which, while representative of current state-of-the-art performance, may not fully capture the diversity in model behavior. Although this does not compromise the strength of our findings, future evaluations with a broader range of models could provide further validation and insights. Additionally, the models used in this study are primarily chat-oriented; reasoning-focused models may exhibit different disambiguation capabilities, and evaluating such models would be a valuable extension. Another constraint is the limited sample size of 300 sentences per language. While this restricts the scale of the analysis, it ensures that each sampling technique operates on an identical and controlled dataset, thereby preserving consistency across evaluations without introducing bias from differing input distributions.

\section*{Acknowledgments}

This paper has been conducted in compliance with the ethical standards of Swansea University. Hough's work is supported by the EPSRC grant EP/X009343/1 `FLUIDITY'. Also, we would like to extend our sincere appreciation to the OpenAI Researcher Access Program for generously providing the credits that made the development of this project possible.

\bibliographystyle{acl_natbib}
\bibliography{ranlp2023}

\begin{thebibliography}{28}
\expandafter\ifx\csname natexlab\endcsname\relax\def\natexlab#1{#1}\fi

\bibitem[{Ahmadnia et~al.(2025)Ahmadnia, Jordehi, Heyran, Mirroshandel, Rambow,
  and Caragea}]{ahmadnia2025active}
Saeed Ahmadnia, Arash~Yousefi Jordehi, Mahsa Hosseini~Khasheh Heyran,
  Seyed~Abolghasem Mirroshandel, Owen Rambow, and Cornelia Caragea. 2025.
\newblock Active few-shot learning for text classification.
\newblock \emph{arXiv preprint arXiv:2502.18782}.

\bibitem[{Aina and Linzen(2021)}]{aina2021language}
Laura Aina and Tal Linzen. 2021.
\newblock The language model understood the prompt was ambiguous: Probing
  syntactic uncertainty through generation.
\newblock \emph{arXiv preprint arXiv:2109.07848}.

\bibitem[{Cahyawijaya et~al.(2024)Cahyawijaya, Zhang, Lovenia, Cruz, Nomoto,
  and Aji}]{cahyawijaya2024thank}
Samuel Cahyawijaya, Ruochen Zhang, Holy Lovenia, Jan Christian~Blaise Cruz,
  Hiroki Nomoto, and Alham~Fikri Aji. 2024.
\newblock Thank you, stingray: Multilingual large language models can not (yet)
  disambiguate cross-lingual word sense.
\newblock \emph{arXiv preprint arXiv:2410.21573}.

\bibitem[{Gu et~al.(2024)Gu, Jiang, Shi, Tan, Zhai, Xu, Li, Shen, Ma, Liu
  et~al.}]{gu2024survey}
Jiawei Gu, Xuhui Jiang, Zhichao Shi, Hexiang Tan, Xuehao Zhai, Chengjin Xu, Wei
  Li, Yinghan Shen, Shengjie Ma, Honghao Liu, et~al. 2024.
\newblock A survey on llm-as-a-judge.
\newblock \emph{arXiv preprint arXiv:2411.15594}.

\bibitem[{Han et~al.(2024)Han, Zhang, Yao, Jin, Xu, and He}]{han2024llm}
Shanshan Han, Qifan Zhang, Yuhang Yao, Weizhao Jin, Zhaozhuo Xu, and Chaoyang
  He. 2024.
\newblock Llm multi-agent systems: Challenges and open problems.
\newblock \emph{arXiv preprint arXiv:2402.03578}.

\bibitem[{Jurgens and Klapaftis(2013)}]{jurgens2013semeval}
David Jurgens and Ioannis Klapaftis. 2013.
\newblock Semeval-2013 task 13: Word sense induction for graded and non-graded
  senses.
\newblock In \emph{Second Joint Conference on Lexical and Computational
  Semantics (* SEM), Volume 2: Proceedings of the Seventh International
  Workshop on Semantic Evaluation (SemEval 2013)}, pages 290--299.

\bibitem[{Kang et~al.(2023)Kang, Blevins, and
  Zettlemoyer}]{kang_translate_2023}
Haoqiang Kang, Terra Blevins, and Luke Zettlemoyer. 2023.
\newblock \href {http://arxiv.org/abs/2304.13803} {Translate to {Disambiguate}:
  {Zero}-shot {Multilingual} {Word} {Sense} {Disambiguation} with {Pretrained}
  {Language} {Models}}.
\newblock ArXiv:2304.13803 [cs].

\bibitem[{Kibria et~al.(2024)Kibria, Dipta, and Adnan}]{kibria2024functional}
Raihan Kibria, Sheikh Dipta, and Muhammad Adnan. 2024.
\newblock On functional competence of llms for linguistic disambiguation.
\newblock In \emph{Proceedings of the 28th Conference on Computational Natural
  Language Learning}, pages 143--160.

\bibitem[{Lai et~al.(2025)Lai, Zhang, Lin, Zhang, Ye, Yan, Xu, He, Wang, Zhang
  et~al.}]{lai2025sas}
Peichao Lai, Kexuan Zhang, Yi~Lin, Linyihan Zhang, Feiyang Ye, Jinhao Yan,
  Yanwei Xu, Conghui He, Yilei Wang, Wentao Zhang, et~al. 2025.
\newblock Sas-bench: A fine-grained benchmark for evaluating short answer
  scoring with large language models.
\newblock \emph{arXiv preprint arXiv:2505.07247}.

\bibitem[{Li et~al.(2024)Li, Wang, Zeng, Wu, and Yang}]{li2024survey}
Xinyi Li, Sai Wang, Siqi Zeng, Yu~Wu, and Yi~Yang. 2024.
\newblock A survey on llm-based multi-agent systems: workflow, infrastructure,
  and challenges.
\newblock \emph{Vicinagearth}, 1(1):9.

\bibitem[{Ma et~al.(2023)Ma, Zhang, Bian, Liu, Zhang, Zhao, Zhang, Fu, Hu, and
  Wu}]{ma2023fairness}
Huan Ma, Changqing Zhang, Yatao Bian, Lemao Liu, Zhirui Zhang, Peilin Zhao, Shu
  Zhang, Huazhu Fu, Qinghua Hu, and Bingzhe Wu. 2023.
\newblock Fairness-guided few-shot prompting for large language models.
\newblock \emph{Advances in Neural Information Processing Systems},
  36:43136--43155.

\bibitem[{Mallen and Belrose(2024)}]{mallen2024balancing}
Alex Mallen and Nora Belrose. 2024.
\newblock Balancing label quantity and quality for scalable elicitation.
\newblock \emph{arXiv preprint arXiv:2410.13215}.

\bibitem[{Manjavacas and Fonteyn(2022)}]{manjavacas2022non}
Enrique Manjavacas and Lauren Fonteyn. 2022.
\newblock Non-parametric word sense disambiguation for historical languages.
\newblock In \emph{Proceedings of the 2nd International Workshop on Natural
  Language Processing for Digital Humanities}, pages 123--134.

\bibitem[{Mann et~al.(2020)Mann, Ryder, Subbiah, Kaplan, Dhariwal, Neelakantan,
  Shyam, Sastry, Askell, Agarwal et~al.}]{mann2020language}
Ben Mann, N~Ryder, M~Subbiah, J~Kaplan, P~Dhariwal, A~Neelakantan, P~Shyam,
  G~Sastry, A~Askell, S~Agarwal, et~al. 2020.
\newblock Language models are few-shot learners.
\newblock \emph{arXiv preprint arXiv:2005.14165}, 1:3.

\bibitem[{Miller(1995)}]{miller1995wordnet}
George~A Miller. 1995.
\newblock Wordnet: a lexical database for english.
\newblock \emph{Communications of the ACM}, 38(11):39--41.

\bibitem[{Minaee et~al.(2024)Minaee, Mikolov, Nikzad, Chenaghlu, Socher,
  Amatriain, and Gao}]{minaee_large_2024}
Shervin Minaee, Tomas Mikolov, Narjes Nikzad, Meysam Chenaghlu, Richard Socher,
  Xavier Amatriain, and Jianfeng Gao. 2024.
\newblock \href {http://arxiv.org/abs/2402.06196} {Large {Language} {Models}:
  {A} {Survey}}.
\newblock ArXiv:2402.06196 [cs].

\bibitem[{Navigli and Ponzetto(2010)}]{navigli2010babelnet}
Roberto Navigli and Simone~Paolo Ponzetto. 2010.
\newblock Babelnet: Building a very large multilingual semantic network.
\newblock In \emph{Proceedings of the 48th annual meeting of the association
  for computational linguistics}, pages 216--225.

\bibitem[{Ortega-Martín et~al.(2023)Ortega-Martín, García-Sierra, Ardoiz,
  Álvarez, Armenteros, and Alonso}]{ortega-martin_linguistic_2023}
Miguel Ortega-Martín, Óscar García-Sierra, Alfonso Ardoiz, Jorge Álvarez,
  Juan~Carlos Armenteros, and Adrián Alonso. 2023.
\newblock \href {http://arxiv.org/abs/2302.06426} {Linguistic ambiguity
  analysis in {ChatGPT}}.
\newblock ArXiv:2302.06426 [cs].

\bibitem[{Pecher et~al.(2024)Pecher, Srba, Bielikova, and
  Vanschoren}]{pecher2024automatic}
Branislav Pecher, Ivan Srba, Maria Bielikova, and Joaquin Vanschoren. 2024.
\newblock Automatic combination of sample selection strategies for few-shot
  learning.
\newblock \emph{arXiv preprint arXiv:2402.03038}.

\bibitem[{Qorib et~al.(2024)Qorib, Moon, and Ng}]{qorib2024decoder}
Muhammad Qorib, Geonsik Moon, and Hwee~Tou Ng. 2024.
\newblock Are decoder-only language models better than encoder-only language
  models in understanding word meaning?
\newblock In \emph{Findings of the Association for Computational Linguistics
  ACL 2024}, pages 16339--16347.

\bibitem[{Sainz et~al.(2023)Sainz, de~Lacalle, Agirre, and
  Rigau}]{sainz_what_2023}
Oscar Sainz, Oier~Lopez de~Lacalle, Eneko Agirre, and German Rigau. 2023.
\newblock \href {https://aclanthology.org/2023.gwc-1.40/} {What do language
  models know about word senses? zero-shot {WSD} with language models and
  domain inventories}.
\newblock In \emph{Proceedings of the 12th Global Wordnet Conference}, pages
  331--342, University of the Basque Country, Donostia - San Sebastian, Basque
  Country. Global Wordnet Association.

\bibitem[{Sumanathilaka et~al.(2024{\natexlab{a}})Sumanathilaka, Micallef, and
  Hough}]{sumanathilaka2024assessing}
Deshan Sumanathilaka, Nicholas Micallef, and Julian Hough. 2024{\natexlab{a}}.
\newblock Assessing gpt's potential for word sense disambiguation: A
  quantitative evaluation on prompt engineering techniques.
\newblock In \emph{2024 IEEE 15th Control and System Graduate Research
  Colloquium (ICSGRC)}, pages 204--209. IEEE.

\bibitem[{Sumanathilaka et~al.(2025{\natexlab{a}})Sumanathilaka, Micallef, and
  Hough}]{sumanathilaka2025exploring}
Deshan Sumanathilaka, Nicholas Micallef, and Julian Hough. 2025{\natexlab{a}}.
\newblock \href {https://doi.org/10.1109/ICNLP65360.2025.11108362} {Exploring
  the impact of temperature on large language models: A case study for
  classification task based on word sense disambiguation}.
\newblock In \emph{2025 7th International Conference on Natural Language
  Processing (ICNLP)}, pages 178--182.

\bibitem[{Sumanathilaka et~al.(2025{\natexlab{b}})Sumanathilaka, Micallef, and
  Hough}]{sumanathilaka_glossgpt_2025}
Deshan Sumanathilaka, Nicholas Micallef, and Julian Hough. 2025{\natexlab{b}}.
\newblock Glossgpt: Gpt for word sense disambiguation using few-shot
  chain-of-thought prompting.
\newblock \emph{Procedia Computer Science}, 257:785--792.

\bibitem[{Sumanathilaka et~al.(2024{\natexlab{b}})Sumanathilaka, Micallef, and
  Hough}]{sumanathilaka2024can}
Deshan~Koshala Sumanathilaka, Nicholas Micallef, and Julian Hough.
  2024{\natexlab{b}}.
\newblock \href {https://aclanthology.org/2024.nlpaics-1.12/} {Can {LLM}s
  assist with ambiguity? a quantitative evaluation of various large language
  models on word sense disambiguation}.
\newblock In \emph{Proceedings of the First International Conference on Natural
  Language Processing and Artificial Intelligence for Cyber Security}, pages
  97--108, Lancaster, UK. International Conference on Natural Language
  Processing and Artificial Intelligence for Cyber Security.

\bibitem[{Vayani et~al.(2025)Vayani, Dissanayake, Watawana, Ahsan, Sasikumar,
  Thawakar, Ademtew, Hmaiti, Kumar, Kukreja et~al.}]{vayani2025all}
Ashmal Vayani, Dinura Dissanayake, Hasindri Watawana, Noor Ahsan, Nevasini
  Sasikumar, Omkar Thawakar, Henok~Biadglign Ademtew, Yahya Hmaiti, Amandeep
  Kumar, Kartik Kukreja, et~al. 2025.
\newblock All languages matter: Evaluating lmms on culturally diverse 100
  languages.
\newblock In \emph{Proceedings of the Computer Vision and Pattern Recognition
  Conference}, pages 19565--19575.

\bibitem[{Yae et~al.(2024)Yae, Skelly, Ranly, and LaCasse}]{yae2024leveraging}
Jung~H Yae, Nolan~C Skelly, Neil~C Ranly, and Phillip~M LaCasse. 2024.
\newblock Leveraging large language models for word sense disambiguation.
\newblock \emph{Neural Computing and Applications}, pages 1--18.

\bibitem[{Yu et~al.(2021)Yu, Zuo, Jiang, Ren, Zhao, and
  Zhang}]{yu_fine-tuning_2021}
Yue Yu, Simiao Zuo, Haoming Jiang, Wendi Ren, Tuo Zhao, and Chao Zhang. 2021.
\newblock \href {http://arxiv.org/abs/2010.07835} {Fine-{Tuning} {Pre}-trained
  {Language} {Model} with {Weak} {Supervision}: {A} {Contrastive}-{Regularized}
  {Self}-{Training} {Approach}}.
\newblock ArXiv:2010.07835 [cs].

\end{thebibliography}


\end{document}